\documentclass[journal, 10pt, twocolumn]{IEEEtran}

\usepackage{cite}
\usepackage{amsmath,amssymb,amsfonts}
\usepackage{algorithmic}
\usepackage{graphicx}
\usepackage{subcaption} 
\usepackage{mwe} 
\usepackage{textcomp}
\usepackage{xcolor}
\usepackage[hidelinks]{hyperref}
\usepackage{booktabs} 
\usepackage{multirow} 
\usepackage{tcolorbox} 
\usepackage{placeins} 

\hypersetup{
    breaklinks=true,
    pdfcreator={LaTeX},
    pdfproducer={pdflatex}
}

\usepackage{tikz}
\usetikzlibrary{positioning,shapes.geometric,shapes.symbols,shapes.misc,arrows}

\AtBeginEnvironment{thebibliography}{\sloppy\setlength{\emergencystretch}{2em}}

\title{Evaluating Synthetic Data for Baggage Trolley Detection in Airport Logistics}

\author{
    \IEEEauthorblockN{
        Abdeldjalil Taibi\IEEEauthorrefmark{1},
        Mohmoud Badlis\IEEEauthorrefmark{1}, 
        Amina Bensalem\IEEEauthorrefmark{1}, 
        Belkacem Zouilekh\IEEEauthorrefmark{2}, and 
        Mohammed Brahimi\IEEEauthorrefmark{1}
    }
    \vspace{0.15cm} 
    
    \IEEEauthorblockA{
        \IEEEauthorrefmark{1}\textit{The National School of Artificial Intelligence (ENSIA)}\\
        Algiers 16000, Algeria\\
        Email: \{abd-eldjalil.taibi, mohmoud.badlis, mohamed.brahimi\}@ensia.edu.dz
    }
    \vspace{0.15cm}
    
    \IEEEauthorblockA{
        \IEEEauthorrefmark{2}\textit{Information Systems and Development Department, Algiers Airport}\\
        Algiers 16000, Algeria\\
        Email: b.zouilekh@aeroportalger.dz
    }
}

\begin{document}

\maketitle

\begin{abstract}

Efficient luggage trolley management is critical for reducing congestion and ensuring asset availability in modern airports. Automated detection systems face two main challenges. First, strict security and privacy regulations limit large-scale data collection. Second, existing public datasets lack the diversity, scale, and annotation quality needed to handle dense, overlapping trolley arrangements typical of real-world operations.

To address these limitations, we introduce a synthetic data generation pipeline based on a high-fidelity Digital Twin of Algiers International Airport using NVIDIA Omniverse. The pipeline produces richly annotated data with oriented bounding boxes, capturing complex trolley formations, including tightly nested chains. We evaluate YOLO-OBB using five training strategies: real-only, synthetic-only, linear probing, full fine-tuning, and mixed training. This allows us to assess how synthetic data can complement limited real-world annotations.

Our results show that mixed training with synthetic data and only 40 percent of real annotations matches or exceeds the full real-data baseline, achieving 0.94 mAP@50 and 0.77 mAP@50-95, while reducing annotation effort by 25 to 35 percent. Multi-seed experiments confirm strong reproducibility with a standard deviation below 0.01 on mAP@50, demonstrating the practical effectiveness of synthetic data for automated trolley detection.

\end{abstract}
\section{Introduction}
\label{sec:intro}

\IEEEPARstart{T}{he} transition towards ``Smart Airports'' relies heavily on computer vision to automate surveillance, passenger flow analysis, and asset management \cite{airport4_0_review}. Among the critical assets in airport logistics, luggage trolleys play a critical role. Efficient trolley management is essential for preventing congestion, ensuring availability for passengers, and maintaining safety standards. However, automating the detection and tracking of these assets presents unique challenges due to the visual complexity of airport terminals, which are characterized by high crowd density, reflective surfaces, and dynamic lighting conditions.

Despite progress in general object detection, there is a significant lack of public datasets designed specifically for airport trolley surveillance. This scarcity is largely driven by the legal, privacy, and security constraints imposed by airport administrations and security authorities, which restrict data acquisition, dissemination, and annotation inside operational terminals. Although open source platforms like Roboflow host community contributions such as the \textit{Airport Trolley} set by DuongBopBi \cite{roboflow_trolley_duong} and the \textit{Trolley} set by KSS \cite{roboflow_trolley_kss}, these datasets remain limited in both size and quality. Specifically, they contain only 197 and 107 frames respectively, which is insufficient for training robust deep learning models. Moreover, they use only Axis Aligned Bounding Boxes (AABB). AABB labels are unsuitable for the complex geometry of airport environments, where trolleys are often oriented diagonally or connected in ``chained'' formations. In such cases, AABB labels capture excessive background and overlap, making it difficult for models to separate individual units in crowded areas.

To address the difficulty of collecting large scale, annotated data in high security airport zones, we propose the integration of synthetic data as a scalable alternative to extensive in terminal recording. Synthetic environments enable the controlled generation of rare and complex scenarios that are difficult to capture in the real world, while avoiding many of the operational and legal barriers that limit the release of real surveillance footage. In this work, we leverage \textbf{NVIDIA Omniverse} to create a simplified ``Digital Twin'' of the Algiers International Airport. The virtual environment models the trolley variants deployed at the airport and generates training samples under varied camera poses, lighting, and crowd levels, with the goal of reducing the amount of real data required to reach high detection performance.

Consequently, this paper presents a comprehensive study on trolley detection using a hybrid data approach. Our main contributions are as follows:

\begin{itemize}
    \item \textbf{A Real World Airport Trolley Dataset:} We introduce a curated dataset collected from public video sources, explicitly selecting scenes that feature challenging conditions such as chained trolleys, heavy occlusion, and diverse viewpoints.
    
    \item \textbf{A Synthetic "Digital Twin" Dataset:} We provide a large scale synthetic dataset generated within a digital replica of the Algiers International Airport, featuring diverse trolley models and distinct points of view (POV) to enhance model learning.
    
    \item \textbf{Systematic Evaluation of Synthetic Data Utility:} 
We conduct controlled experiments comparing multiple training strategies: 
real only training, synthetic only training, synthetic pretraining followed by full fine tuning, synthetic pretraining with prediction head only finetuning, and mixed training using both real and synthetic data. 
This enables us to quantify when and how synthetic data improves detection performance and to what extent it reduces the required amount of real world training data.

    \item \textbf{Public Benchmark Release:} To facilitate future research in airport asset management, we release both the annotated datasets and the trained model benchmarks to the community.
\end{itemize}
\section{Related Work}
\label{sec:related}

\IEEEPARstart{T}{his} section reviews existing airport computer vision methods, commercial systems, and object counting approaches. We also discuss why existing datasets are not enough for this problem and how synthetic data can help.

\subsection{Computer Vision in Airport Logistics}
Computer vision is widely used in airports, but most research focuses on security or outdoor operations. For example, recent studies have looked at detecting debris on runways using drones \cite{alkuwari2023uav} or tracking fuel trucks to avoid accidents \cite{goulart2022apron}.  Inside the terminal, most work deals with finding abandoned luggage \cite{vrsalovic2023abandoned} or monitoring passenger queues using vision-based queue length estimation from surveillance streams \cite{zhao2020realtimequeue}.

This evident focus on security and safety reinforces the importance of researching \textit{operational logistics}, specifically the domain of trolley management. Unlike static abandoned bags or distinct service vehicles, trolleys are dynamic assets, frequently moved by passengers and often clustered in overlapping chains. By addressing this overlooked area, we aim to replace inefficient manual monitoring with an automated system that prioritizes asset availability and operational flow.

\subsection{Commercial and Industrial Solutions}
In the industry, some companies have built systems to track trolleys. A leading example is the system by Wanzl and NAITEC, which combines cameras with RFID or Bluetooth tags attached to the trolleys \cite{WanzlNAITEC2023}. These systems can tell staff exactly where trolleys are and if a station is empty.

While these systems work well, they are expensive because they require a combination of physical tags on every trolley and dedicated infrastructure across the terminal, particularly the corresponding readers and antennas needed to reliably detect those tags. If a tag or a reader/antenna breaks, the tracking fails. Our project proposes a \textbf{pure computer vision solution} that uses standard security cameras. This removes the need for tag hardware and its supporting reader/antenna network, making the system easier to deploy in existing airports.

\subsection{Approaches to Object Counting: Detection vs. Density}
To manage trolleys automatically, the system needs to count them accurately. There are two main ways to do this in the literature:

\subsubsection{Density Based and Point Based Methods}
When objects are very crowded, researchers traditionally use "density estimation." Methods like DMCount \cite{Wang2020DMCount} create a heat map of the image to estimate the total number of people or objects. Newer baseline methods like P2PNet \cite{Song2021P2PNet} predict a point for the center of each object instead of a box.

Recent comprehensive surveys from 2025 and 2026 emphasize that density estimation has evolved significantly, incorporating vision transformers (ViTs), diffusion models, and multi-scale attention mechanisms to handle extreme scale variations in dense crowds \cite{afik2025crowd, chen2025comprehensive, elsepae2026deep}. Contemporary architectures, such as block based density map generators with channel wise attention \cite{elharrouss2024crowd} and joint density detection frameworks like DEO Net \cite{phan2024deo}, have pushed the boundaries of accuracy in highly congested scenes. Furthermore, recent semi supervised approaches are actively attempting to reduce the heavy reliance on fine grained point annotations \cite{count2density2025}.

Despite these advanced deep learning paradigms, the fundamental limitation of these approaches remains their inherent design for crowd counting, where the target object (typically a human head) is compact and approximates a single point. In contrast, luggage trolleys exhibit a large spatial context with complex geometry. When applied to such objects, density based models frequently suffer from \textit{fragmented activation}, where separate trolley components (handle, wheels, basket) are detected as independent instances. This leads to significant overcounting and fails to treat the trolley as a singular, cohesive unit.

\subsubsection{Detection Based Methods (YOLO)}
The second approach is "detection," where the model draws a box around each object. The \textbf{YOLO (You Only Look Once)} family \cite{Redmon2016YOLO} is the standard choice for this because it is fast enough to run on live video. The problem with standard YOLO is that it uses \textbf{Axes Aligned Bounding Boxes (AABB)}. When trolleys are pushed together in a chain, these \textbf{AABB} overlap heavily, making it hard for the model to see them as separate objects.

We improve on this by using \textbf{Oriented Bounding Boxes (OBB)}. This allows the model to draw a rotated box that fits the trolley perfectly, separating the nested units even in a tight chain.

\subsection{Sim2Real Transfer, Domain Adaptation, and Oriented Detection}
A major challenge in training robust object detectors for restricted environments like airports is the scarcity of annotated real world data. To bridge the resulting "reality gap" between simulation and the real world, researchers increasingly rely on Sim2Real transfer \cite{tremblay2018training}. Modern synthetic data generation pipelines, such as BlenderProc \cite{denninger2019blenderproc} and NVIDIA Isaac Sim \cite{makoviychuk2021isaac}, have demonstrated that photorealistic rendering and domain randomization improve generalization to real indoor environments. While Conde et al. \cite{conde2023generation} successfully applied these tools to track solid objects (e.g., vehicles) in airports, applying these techniques to complex, semi transparent wiremesh structures like luggage trolleys remains under explored. Our custom "Digital Twin" of the Algiers airport builds upon these simulation foundations, focusing heavily on the unique geometric challenges of nested, transparent objects.

Furthermore, applying models trained entirely on synthetic data to a real world target domain inevitably causes a performance drop due to distribution shifts. To mitigate this, Unsupervised Domain Adaptation (UDA) methods have been extensively studied. Foundational feature level approaches like Deep CORAL \cite{sun2016deep} address this by aligning the second order statistics (covariances) of source and target feature distributions. Alternatively, adversarial approaches such as Domain Adversarial Neural Networks (DANN) \cite{ganin2016domain} employ gradient reversal to learn features that are discriminative for the main task but statistically invariant across domains. At the pixel level, image to image translation frameworks like CycleGAN \cite{zhu2017unpaired} are highly effective for synthetic to real transfer, mapping synthetic source images directly into the visual style of the target domain using unpaired data. While our current methodology establishes a strong adaptation baseline via mixed data training, evaluating these advanced UDA frameworks particularly adversarial alignment and CycleGAN based stylistic translation represents a natural evolution for our future work.

Finally, solving the reality gap for this specific asset requires specialized detection architectures. Standard horizontal boxes (AABB) fail for densely clustered chains of trolleys due to excessive overlap. Consequently, recent literature has advanced Oriented Bounding Box (OBB) detection significantly. Beyond the YOLO family \cite{Redmon2016YOLO}, researchers have developed highly accurate models for rotated objects: the RoI Transformer \cite{ding2019learning} applies spatial transformations to region proposals to learn rotation parameters; Oriented R-CNN \cite{xie2021oriented} introduces a lightweight oriented Region Proposal Network; and S2ANet \cite{han2021align} uses deep feature alignment to resolve inconsistencies between classification and localization. By natively incorporating OBB paradigms within our Sim2Real pipeline, we leverage these geometric insights to accurately parse severely overlapping trolley instances, separating our approach from traditional crowd counting or basic AABB detection methods.

\subsection{Limitations of Existing Public Datasets and Cross Dataset Adaptation}
There are very few public datasets for airport trolleys. Community datasets like the \textit{Airport Trolley} set \cite{roboflow_trolley_duong} and \textit{KSS Trolley} set \cite{roboflow_trolley_kss} exist, but they present two primary domain gaps when applied to surveillance environments:
\begin{enumerate}
    \item \textbf{Viewpoint Discrepancy:} These datasets consist primarily of ground level footage captured via mobile phones. This contrasts sharply with the high angle, ceiling mounted perspective of airport security cameras, creating a significant spatial distribution shift.
    \item \textbf{Annotation Incompatibility:} The labels rely entirely on horizontal bounding boxes (AABB), which, as previously discussed, suffer from excessive overlap when applied to densely clustered or chained trolleys.
\end{enumerate}

To evaluate these limitations, we conducted preliminary cross dataset training at the inception of this study. Initial experiments training exclusively on these public AABB datasets resulted in severely degraded detection metrics. This confirmed that the viewpoint and label type gaps restrict direct generalizability. 

However, rather than discarding these datasets entirely, we recognized the value of the diverse trolley appearances they contained. To utilize them, we programmatically converted their AABB annotations into an Oriented Bounding Box (OBB) format and incorporated this adapted data into our custom training pool. This allowed us to expand the morphological diversity of our dataset while maintaining the strict OBB geometry required for our detection architecture.

Ultimately, we address the remaining data scarcity by introducing a new hybrid dataset. By combining synthetic overhead CCTV imagery with these curated real world samples, we overcome data restrictions and bridge the domain gap. Furthermore, all data is annotated with Oriented Bounding Boxes (OBB) to accurately detect complex, overlapping chains of nested trolleys.
\section{Dataset Description}
\label{sec:dataset}

\IEEEPARstart{T}{o} address the limitations of existing public datasets, namely the inability to handle densely chained trolleys and complex occlusions, we constructed a hybrid dataset comprising real-world footage and a synthetic "Digital Twin" of the Algiers International Airport. Table \ref{tab:dataset_stats} summarizes the dataset distribution.

\subsection{Real World Airport Trolley Dataset}
The real-world component consists of \textbf{1,504 frames} curated to represent the challenging visual conditions of operational airport environments. To ensure a robust evaluation of density handling, we extensively annotated the dataset, resulting in a total of \textbf{14,080 oriented bounding boxes}. On average, each frame contains approximately \textbf{10.17 trolley instances}, a density significantly higher than standard public datasets. This density tests the detector’s ability to separate nested and overlapping trolley chains.

\subsubsection{Data Collection}
Due to strict security regulations preventing direct access to the airport's internal CCTV network, we adopted a multi-source collection strategy:
\begin{itemize}
    \item \textbf{Public Repositories:} We curated high-quality footage from public YouTube walking tours (e.g., Algiers Airport arrivals and departure halls), selecting segments that capture trolleys in natural motion.
   \item \textbf{On-Site Collection:} To ensure operational realism while strictly complying with privacy protocols, a 200-frame dataset was captured at Algiers International Airport using handheld devices to mimic security patrol conditions. This footage was exclusively collected, anonymized, annotated, and officially authorized for our academic use by the airport's IT department.
\end{itemize}

\subsubsection{Diversity and Challenges}
The dataset covers a wide range of viewpoints, including side profile and rear/front views. Special emphasis was placed on capturing "real-world noise" that synthetic data often lacks, including:
\begin{itemize}
    \item \textbf{Motion Blur:} Fast-moving trolleys in the hands of rushing passengers.
    \item \textbf{Occlusion:} Severe visual obstruction caused by luggage piles, crowd density, and structural pillars.
    \item \textbf{Lighting Variance:} Scenes ranging from naturally lit atrium areas to artificially lit corridors.
\end{itemize}

\begin{figure}[htbp]
    \centering
    \includegraphics[width=0.48\textwidth]{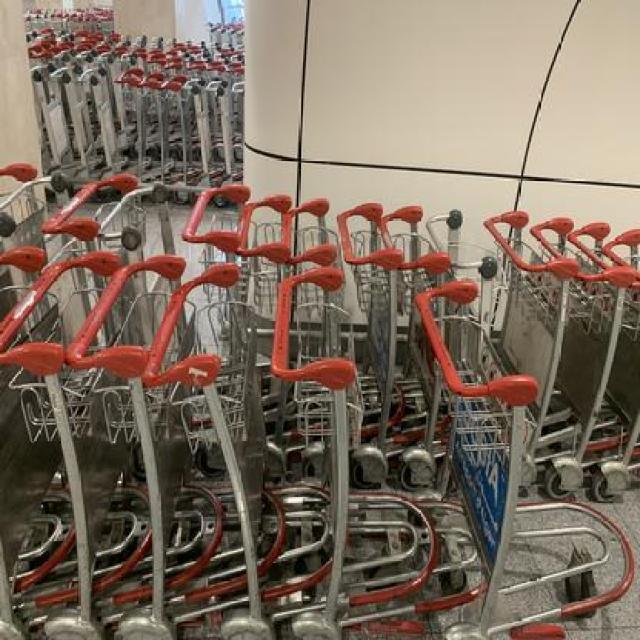}
    \caption{Real-world dataset sample 1.}
    \label{fig:real_sample_1}
\end{figure}

\begin{figure}[htbp]
    \centering
    \includegraphics[width=0.48\textwidth]{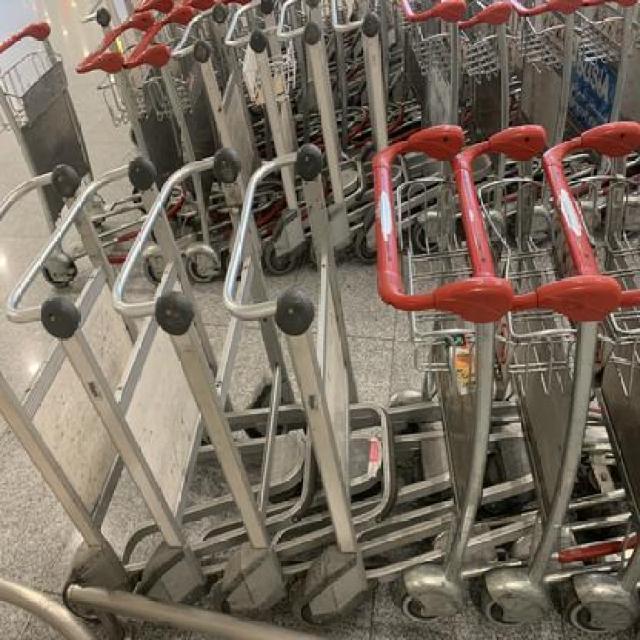}
    \caption{Real-world dataset sample 2.}
    \label{fig:real_sample_2}
\end{figure}

\subsection{Synthetic Dataset: The Algiers Digital Twin}
To overcome the scarcity of diverse, large-scale annotated data for these complex scenarios, we developed a high-fidelity synthetic dataset using NVIDIA Omniverse. Unlike generic synthetic datasets, this environment functions as a "Digital Twin," specifically modeling the geometry and layout of the Algiers Airport to minimize the simulation-to-reality domain gap.

\subsubsection{Environment and Assets}
We modeled three critical zones where trolley management is most required: the \textit{Arrival Zone}, the \textit{Aerogare} (the transition buffer between exterior and interior), and the \textit{Exterior OpenSpace}.
To ensure realistic feature learning, we created 3D assets of the two specific trolley variants used in Algiers airport: the standard metallic structure with \textbf{grey trims} and the variant with \textbf{red trims}.

\subsubsection{Scenario Randomization and Statistics}
We generated a total of \textbf{817 frames} by randomizing scene parameters to cover edge cases rarely seen in real data. This generation process yielded a massive set of \textbf{8616 annotated bounding boxes}. With an average density of \textbf{10.55 objects per image}, the synthetic set closely mimics the high-density clustering found in our real-world samples, ensuring the model learns to separate deeply nested objects. The generation pipeline focused on:

\begin{itemize}
    \item \textbf{Crowd and Interaction:} We simulated scenarios ranging from single trolleys to complex "chained" formations (nested groups of 12 to 18 units). Scenarios also included trolleys loaded with diverse luggage and dynamic human interactions (passengers pushing trolleys or walking past to create occlusion).
    \item \textbf{Camera Placement:} To match the real-world data collection and simulate operational ground-level inspections, cameras were placed at a \textit{Phone-level View (0.8-1.2m)} to capture side profiles, seamlessly mimicking patrol robots or handheld staff inspections.
    \item \textbf{Entry Dynamics:} Trolleys were programmed to enter scenes at varying trajectories (vertical, horizontal, and $45^{\circ}$ angles) through automatic doors to diversify the geometric projection of the objects.
\end{itemize}

\begin{figure}[htbp]
    \centering
    \includegraphics[width=0.48\textwidth]{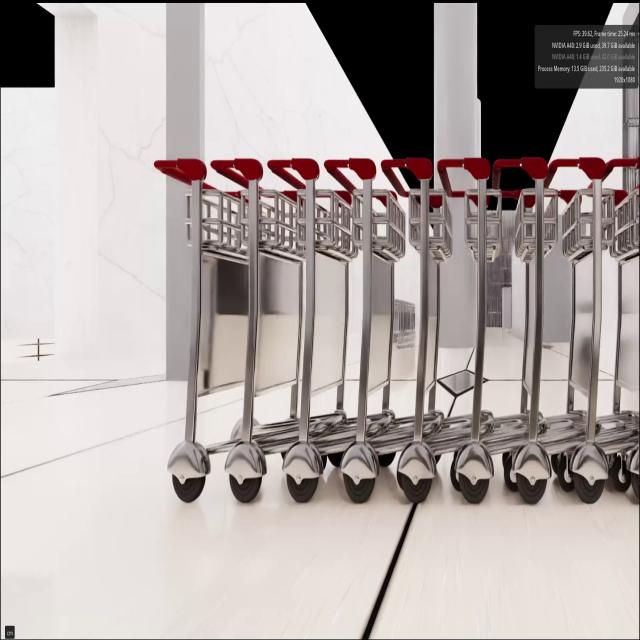}
    \caption{Synthetic dataset sample from the Algiers Digital Twin.}
    \label{fig:synth_sample_1}
\end{figure}

\begin{figure}[htbp]
    \centering
    \includegraphics[width=0.48\textwidth]{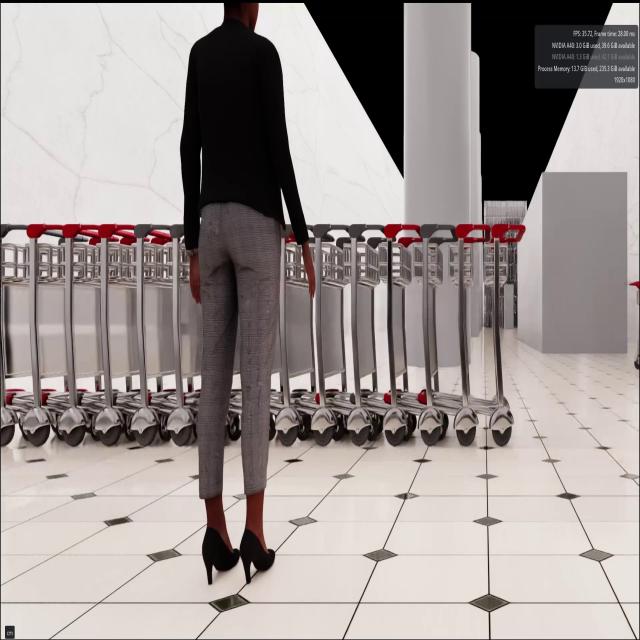}
    \caption{Synthetic dataset sample 2 from the Algiers Digital Twin.}
    \label{fig:synth_sample_2}
\end{figure}

\subsection{Annotation Pipeline}
\begin{figure}[htbp]
    \centering
    \includegraphics[width=0.5\textwidth]{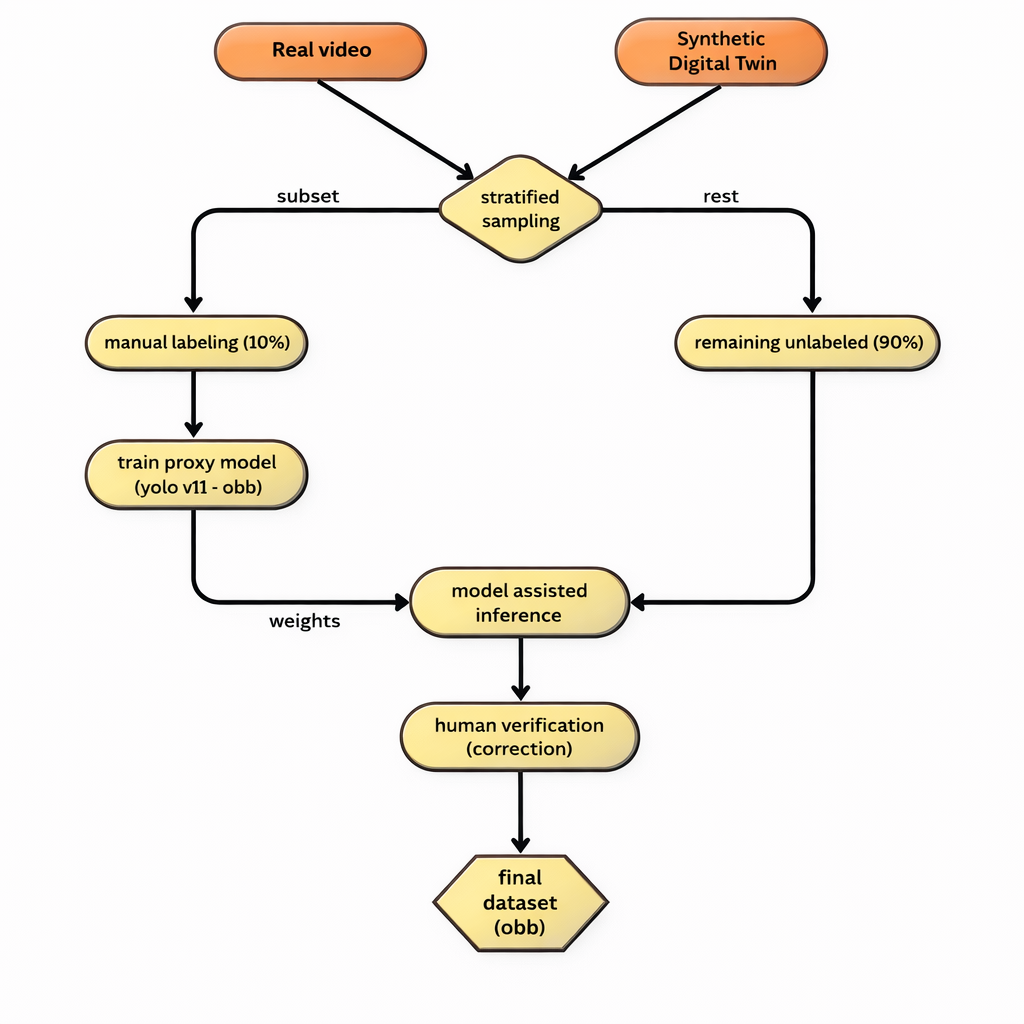}
    \caption{The semi-automated ``Human-in-the-Loop'' annotation workflow.}
    \label{fig:workflow}
\end{figure}

To ensure high-quality supervision while maintaining scalability, we employed a semi-automated ``human-in-the-loop'' annotation pipeline (Figure \ref{fig:workflow}). The core idea is to manually label a small but representative subset of the video frames that we will annotate, train a lightweight detector on it, use that detector to pre-label the remaining data, and then manually correct all predicted labels to obtain consistent, high-quality ground truth.

\subsubsection{Why Oriented Bounding Boxes (OBB)?}
We explicitly chose the \textbf{YOLO-obb} format (normalized vertices) over standard Axis Aligned Bounding Boxes (AABB). In airport logistics, trolleys are frequently nested in curved chains or oriented diagonally. In such configurations, AABB introduces excessive background noise and high Intersection over Union (IoU) overlap between adjacent units. OBB aligns with the object's principal axis, effectively decoupling individual trolleys in dense clusters and providing the geometric precision required for robotic grasping or counting.

\subsubsection{Semi-Automated Workflow}
Our annotation procedure was designed to maximize efficiency without compromising ground truth quality. Concretely, we followed four steps:

\begin{enumerate}
    \item \textbf{Stratified Sampling (10\%):} For each video sequence, we sampled approximately 10\% of frames across the full timeline (beginning/middle/end) and across different density levels (single trolleys to chained formations). This leverages temporal coherence in video: once a viewpoint and scene are covered, adjacent frames add limited new variation.
    
    \item \textbf{Proxy Model Training:} We trained a lightweight \textbf{YOLO11-obb} model on this manual subset. This model served strictly as a temporary annotation assistant and is architecturally distinct from the heavier models used in our final evaluation to prevent data leakage.
    
    \item \textbf{Model-Assisted Pre-Labeling:} The proxy model generated initial OBB predictions for the remaining 90\% of frames, producing ``draft'' labels (class + oriented box vertices) that accelerate annotation.
    
    \item \textbf{Manual Review and Correction:} All draft labels (for both real and synthetic data) were manually audited in Roboflow. Annotators corrected box orientation/extent, added missed trolleys (false negatives), removed spurious detections (false positives), and ensured consistent ``tightness'' of the OBBs across domains. This final pass is critical to avoid learning domain-specific annotation artifacts and to keep IoU-based evaluation meaningful.
\end{enumerate}

\begin{table}[htbp]
\caption{Dataset Statistics}
\label{tab:dataset_stats}
\centering
\small
\setlength{\tabcolsep}{1.5pt}

\resizebox{\columnwidth}{!}{%
\begin{tabular}{lcccc}
\toprule
\textbf{Subset} & \textbf{Frames} & \textbf{Boxes} & \textbf{Density} & \textbf{Viewpoints} \\
\midrule
Real World & 1,504 & 14,080 & 10.17 & Phone view \\
Synthetic  & 817 & 8616 & 10.55  & Phone view \\
\midrule
\textbf{Total} & \textbf{2321} & \textbf{22696} & \textbf{10.36} & - \\
\bottomrule

\end{tabular}}
\end{table}
\section{Methodology}
\label{sec:methodology}

\subsection{Focus and Scope}
This study evaluates the extent to which synthetic data can reduce the need for extensive real-world data collection and manual labeling for trolley detection in airport environments. Acquiring large-scale, high-quality training data in such settings is often cost-prohibitive and constrained by strict security regulations and privacy laws. 

Consequently, we focus on quantifying whether ``Digital Twin'' data can bridge this gap while maintaining high operational reliability. Specifically, we prioritize \textbf{Precision (avoiding false detections)} and \textbf{Recall (finding all trolleys)}, as these are critical for logistic efficiency, alongside standard detection metrics \textbf{(mAP50 and mAP50-95)} which measure the overall localization quality.

\subsection{Detection Model Selection}
Given the geometric constraints of trolley management specifically the need to disentangle nested chains and overlapping units we utilize the \textbf{YOLO26-obb (Oriented Bounding Box)} architecture. 
Unlike standard axis-aligned detectors, the OBB regression head predicts an additional angular parameter $\theta$, allowing the model to capture diagonal orientations without excessive background inclusion or high IoU overlap between adjacent objects.

\subsection{Training Strategies and Evaluation Protocol}
To ensure a consistent benchmark across all experiments, we curated a dedicated \textbf{Test Set} comprising approximately \textbf{200 frames}, which we manually labeled. This set features an average density of $\sim7$ boxes per frame and serves as the unbiased ground truth for all evaluations.

We designed five distinct training protocols to isolate the mechanisms of synthetic data adaptation:

\begin{enumerate}
    \item \textbf{Real Only Baseline (Gold Standard):} 
    The model is trained exclusively on the full real-world training set ($100\%$ Real). This establishes the upper bound of performance achievable with our current data collection.
    
    \item \textbf{Synthetic Only (Simulation to Reality):} 
    The model is trained exclusively on data from the Algiers Digital Twin and evaluated directly on the real-world test set. This measures the raw \textbf{Simulation to Reality} domain gap and the generalization capability of the synthetic priors.
    
    \item \textbf{Strategy A: Linear Probing (Frozen Backbone):} 
    The model is first pretrained on the full synthetic dataset until convergence. Subsequently, the backbone layers (feature extractor) are \textbf{frozen} (freeze parameter = 10), and only the prediction head is finetuned using incremental subsets of real-world data. This tests the transferability of synthetic features without modifying the core feature extractor.
    
    \item \textbf{Strategy B: Full Fine Tuning (Unfrozen):} 
    Similar to Strategy A, the model is pretrained on synthetic data. However, in this phase, \textbf{all layers are unfrozen} (freeze parameter = 0), allowing the gradients to update the entire network from the backbone to the head on the real-world subsets. This evaluates whether allowing the model to adjust its deep feature representations to the real domain yields better adaptation than head-only training.
    
    \item \textbf{Strategy C: Mixed Training (Data Augmentation):} 
    The model is trained from scratch using a combined dataset containing the full synthetic set plus incremental subsets of real-world data (starting from 5\%, then 10\%, 20\%, etc.). 
    We increased the real data fraction until performance convergence was observed.

    \item \textbf{Matched Real Data Baseline Controls:}
    For each real data subset percentage used in Strategies A, B, and C, we trained an additional \textit{Real Only} model using that exact same real data fraction. This enables a direct subset-matched comparison to verify whether synthetic data acts as an effective domain-specific augmentation and whether pretraining provides a tangible benefit in low-data regimes.
\end{enumerate}

For the three adaptation strategies (Linear Probing, Full FineTuning, and Mixed Training), we increased the real data fraction only until performance on the held-out test set converged (i.e., additional real data produced negligible gains). Therefore, we did not necessarily train these strategies up to 100\% of the real-world dataset. Instead, we stopped once the evaluation metrics plateaued to avoid unnecessary training on larger subsets without measurable improvement. 

Throughout the iterative experiments (Strategies A, B, and C), the percentages denote the fraction of the total available \textbf{real-world} dataset, which consists of \textbf{1,305 images}. The synthetic dataset remains fixed at \textbf{817 images}. Consequently, in Mixed Training (Strategy C), the actual proportion of real-world images the model encounters is significantly lower than the subset percentage suggests. Table \ref{tab:data_splits} details the exact frame counts and effective proportions for each experimental split.

\begin{table}[htbp]
\caption{Dataset Composition Across Experimental Splits}
\label{tab:data_splits}
\centering
\small
\resizebox{\columnwidth}{!}{%
\begin{tabular}{ccccc}
\toprule
\textbf{\shortstack{Real Data\\Subset (\%)}} & \textbf{\shortstack{Real Frames\\($N$)}} & \textbf{\shortstack{Synthetic\\Frames}} & \textbf{\shortstack{Combined Total\\(Strategy C)}} & \textbf{\shortstack{Effective Real \%\\in Mixed Batch}} \\
\midrule
5\%  & 65  & 817 & 882  & $\sim$7.4\%  \\
10\% & 131 & 817 & 948  & $\sim$13.8\% \\
20\% & 261 & 817 & 1078 & $\sim$24.2\% \\
30\% & 392 & 817 & 1209 & $\sim$32.4\% \\
40\% & 522 & 817 & 1339 & $\sim$39.0\% \\
50\% & 653 & 817 & 1470 & $\sim$44.4\% \\
\bottomrule
\end{tabular}
}
\end{table}

\begin{figure}[t]
    \centering
    \includegraphics[width=\columnwidth]{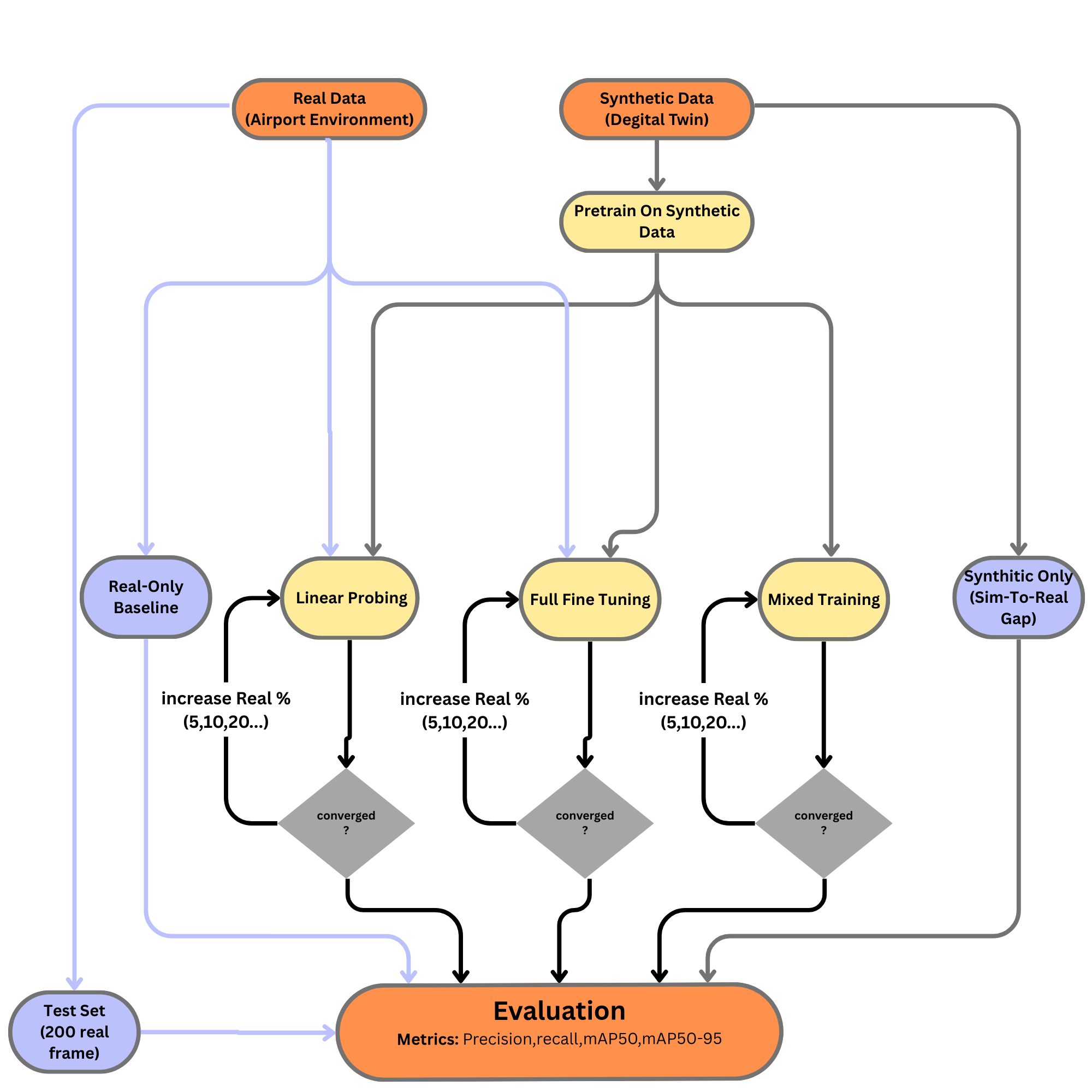}
    \caption{Flowchart of the experimental methodology, illustrating the five distinct training protocols.}
    \label{fig:methodology_flowchart}
\end{figure}
\section{Experimental Setup}
\label{sec:setup}

\subsection{Dataset Splits}
To ensure robust evaluation, we partitioned the data as shown in Table \ref{tab:setup_data_splits}. Note that the synthetic component utilizes \textbf{Phone View} generations to perfectly align with the handheld nature of the real-world operational footage.

\begin{table}[htbp]
\caption{Dataset Partitioning and Source Breakdown}
\label{tab:setup_data_splits}
\centering
\small
\resizebox{\columnwidth}{!}{%
\begin{tabular}{llcl}
\toprule
\textbf{Split} & \textbf{Source} & \textbf{Frames} & \textbf{Notes} \\
\midrule
\multirow{2}{*}{\textbf{Training (80\%)}} & Real-World & 1,044 & Operational footage \\
 & Synthetic & 654 & \textbf{Phone View} \\
\midrule
\multirow{2}{*}{\textbf{Validation (20\%)}} & Real-World & 260 & Sampling from it at matching \% (e.g., 5-50\%) for early stopping \\
 & Synthetic & 163 & 100\% utilized across all mixed experiments \\
\midrule
\textbf{Testing} & Real-World & 200 & \textbf{Strictly Held-Out} (Manual annotation) \\
\bottomrule
\end{tabular}}
\end{table}

\subsection{Evaluation Metrics}
We employ standard Oriented Object Detection metrics. Since our primary objective is \textit{counting accuracy} (i.e., detecting the correct number of trolleys), we prioritize \textbf{Recall} and \textbf{Precision} over fine-grained bounding box localization quality.

\begin{itemize}
    \item \textbf{Recall (Operational Priority):} Measures the percentage of total trolleys successfully detected. In our context, this is the most critical metric, as missing a trolley (False Negative) leads to operational shortages.
    \item \textbf{Precision (Operational Priority):} Measures the accuracy of positive predictions. High precision is essential to prevent false alarms (False Positives) that would waste staff time.
    \item \textbf{mAP50:} Mean Average Precision at IoU threshold 0.50. Used to benchmark general detection capability.
    \item \textbf{mAP50-95:} A stricter measure of localization accuracy, averaging performance across IoU thresholds from 0.50 to 0.95.
\end{itemize}

\subsection{Implementation Details}
Experiments were conducted using the Ultralytics \textbf{YOLO26-obb} framework. We employed the \textbf{AdamW} optimizer with a momentum of \textbf{0.937} and weight decay of \textbf{0.0005} across all runs.

All experiments were trained on computing infrastructure provided by the National School of Artificial Intelligence, equipped with a single NVIDIA H100 GPU (22~GB VRAM), 32~GB system RAM, and approximately 120~GB of available local storage. The CPU model was not exposed by the shared infrastructure; however, we were allocated 25 CPU cores for data loading and training orchestration. The software stack (OS, drivers, and deep learning libraries) followed the most recent available releases at the time of experimentation (Ubuntu-based environment).

To ensure fair comparison, specific hyperparameters were adjusted based on the training protocol. For the \textit{Linear Probing} (Strategy A) and \textit{Full Fine-Tuning} (Strategy B) phases, we utilized a reduced learning rate and shorter warmup period to preserve the pretrained synthetic features. In contrast, the \textit{Mixed Training} (Strategy C) and baseline models (Real-Only/Synthetic-Only) were trained from scratch using standard convergence settings. Table \ref{tab:hyperparams} details these specific configurations.

\begin{table}[htbp]
\caption{Training Hyperparameters: Comparison of Strategies}
\label{tab:hyperparams}
\centering
\small
\resizebox{\columnwidth}{!}{%
\begin{tabular}{lccc}
\toprule
\textbf{Parameter} & \textbf{\shortstack{Linear Probing\\(Strategy A)}} & \textbf{\shortstack{Full Fine-Tuning\\(Strategy B)}} & \textbf{\shortstack{Scratch / Mixed\\(Strategy C)}} \\
\midrule
\textbf{Pre-trained Weights} & Synthetic & Synthetic & None (Random Init) \\
\textbf{Backbone Freezing} & \textbf{First 10 Layers} & \textbf{0 (Unfrozen)} & None \\
\textbf{Initial LR ($lr_0$)} & 0.0005 & 0.0005 & 0.001 \\
\textbf{Warmup Epochs} & 1.0 & 1.0 & 3.0 \\
\midrule
\textbf{Epochs} & \multicolumn{3}{c}{100 (Patience: 25)} \\
\textbf{Image Size} & \multicolumn{3}{c}{640 $\times$ 640} \\
\textbf{Batch Size} & \multicolumn{3}{c}{12} \\
\textbf{Optimizer} & \multicolumn{3}{c}{AdamW} \\
\midrule
\textbf{Augmentations} & \multicolumn{3}{c}{Identical Configuration} \\
\textit{Mosaic} & \multicolumn{3}{c}{1.0} \\
\textit{Mixup} & \multicolumn{3}{c}{0.1} \\
\textit{HSV ($h, s, v$)} & \multicolumn{3}{c}{0.015, 0.7, 0.4} \\
\textit{Translation / Scale} & \multicolumn{3}{c}{0.1 / 0.9} \\
\textit{Degrees / FlipLR} & \multicolumn{3}{c}{$\pm 10.0^{\circ}$ / 0.5} \\
\bottomrule
\end{tabular}}
\end{table}
\section{Results}
\label{sec:results}

\IEEEPARstart{T}{his} section presents the \textbf{quantitative} and \textbf{qualitative} evaluation of the proposed training strategies on the held-out real-world test set (200 frames). We report Precision and Recall, alongside Mean Average Precision (mAP) at IoU 0.50 and 0.50-95.

\subsection{Baseline Performance Bounds}
Table \ref{tab:baselines} establishes the performance boundaries. The \textit{Real-Only (100\%)} model represents the upper bound using all available target data, while the \textit{Synthetic-Only} model quantifies the zero-shot transfer performance.

\begin{table}[htbp]
\caption{Baseline Performance Bounds (Test Set)}
\label{tab:baselines}
\begin{center}
\begin{tabular}{lcccc}
\toprule
\textbf{Model Variant} & \textbf{mAP50} & \textbf{mAP50-95} & \textbf{Precision} & \textbf{Recall} \\
\midrule
Real-Only (100\% Data) & 0.9424 & 0.8010 & 0.9375 & 0.9057 \\
Synthetic-Only (Zero-Shot) & 0.4162 & 0.2184 & 0.4501 & 0.4835 \\
\bottomrule
\end{tabular}
\end{center}
\end{table}

\subsection{Comparison of Training Strategies}
Table \ref{tab:strategy_comparison} provides a comprehensive comparison of the three synthetic adaptation strategies across increasing subsets of real-world data.

The results demonstrate a clear hierarchy of performance. \textbf{Strategy A (Linear Probing)}, where the backbone is frozen, yields the lowest performance, indicating that the synthetic features alone are insufficient for the specific textures of the real domain. \textbf{Strategy B (Full Fine-Tuning)} significantly improves upon this by allowing the backbone to adapt, achieving mAP scores competitive with Mixed Training at higher data percentages (e.g., 50\%). However, \textbf{Strategy C (Mixed Training)} remains the most robust approach in low-data regimes (5-30\%), consistently delivering the highest Recall and mAP.

\begin{table*}[t]
\caption{Comparative Results: Linear Probing vs. Full Fine-Tuning vs. Mixed Training}
\label{tab:strategy_comparison}
\centering
\small
\resizebox{\textwidth}{!}{%
\begin{tabular}{c|cccc|cccc|cccc}
\toprule
\multirow{2}{*}{\textbf{Real Data \%}} & \multicolumn{4}{c|}{\textbf{Strategy A: Linear Probing (Frozen)}} & \multicolumn{4}{c|}{\textbf{Strategy B: Full Fine-Tuning (Unfrozen)}} & \multicolumn{4}{c}{\textbf{Strategy C: Mixed Training}} \\
 & \textbf{mAP50} & \textbf{mAP50-95} & \textbf{Prec.} & \textbf{Recall} & \textbf{mAP50} & \textbf{mAP50-95} & \textbf{Prec.} & \textbf{Recall} & \textbf{mAP50} & \textbf{mAP50-95} & \textbf{Prec.} & \textbf{Recall} \\
\midrule
5\%  & 0.8977 & 0.5455 & 0.8747 & 0.8048 & 0.9140 & 0.5953 & 0.8725 & 0.8291 & \textbf{0.9281} & \textbf{0.7158} & \textbf{0.9017} & \textbf{0.8817} \\
10\% & 0.8515 & 0.4364 & 0.8586 & 0.7560 & 0.9113 & 0.6261 & 0.8725 & 0.8478 & \textbf{0.9326} & \textbf{0.6740} & \textbf{0.9088} & \textbf{0.8782} \\
20\% & 0.8910 & 0.5895 & 0.9034 & 0.7874 & \textbf{0.9250} & 0.6027 & 0.8812 & 0.8507 & 0.9071 & \textbf{0.6593} & \textbf{0.8982} & 0.8280 \\
30\% & 0.8691 & 0.5352 & 0.8376 & 0.8082 & 0.9210 & 0.6274 & 0.8712 & 0.8726 & \textbf{0.9320} & \textbf{0.7341} & \textbf{0.9133} & \textbf{0.8896} \\
40\% & 0.9006 & 0.5934 & 0.8671 & 0.8229 & 0.9215 & 0.6784 & 0.8909 & 0.8712 & \textbf{0.9402} & \textbf{0.7301} & \textbf{0.9141} & \textbf{0.8916} \\
50\% & 0.8997 & 0.6309 & 0.8483 & 0.8387 & \textbf{0.9430} & 0.7268 & 0.8890 & \textbf{0.9110} & 0.9366 & \textbf{0.7655} & \textbf{0.9157} & 0.8846 \\
\bottomrule
\end{tabular}}
\end{table*}

\begin{figure}[htbp]
    \centering
    \includegraphics[width=\columnwidth]{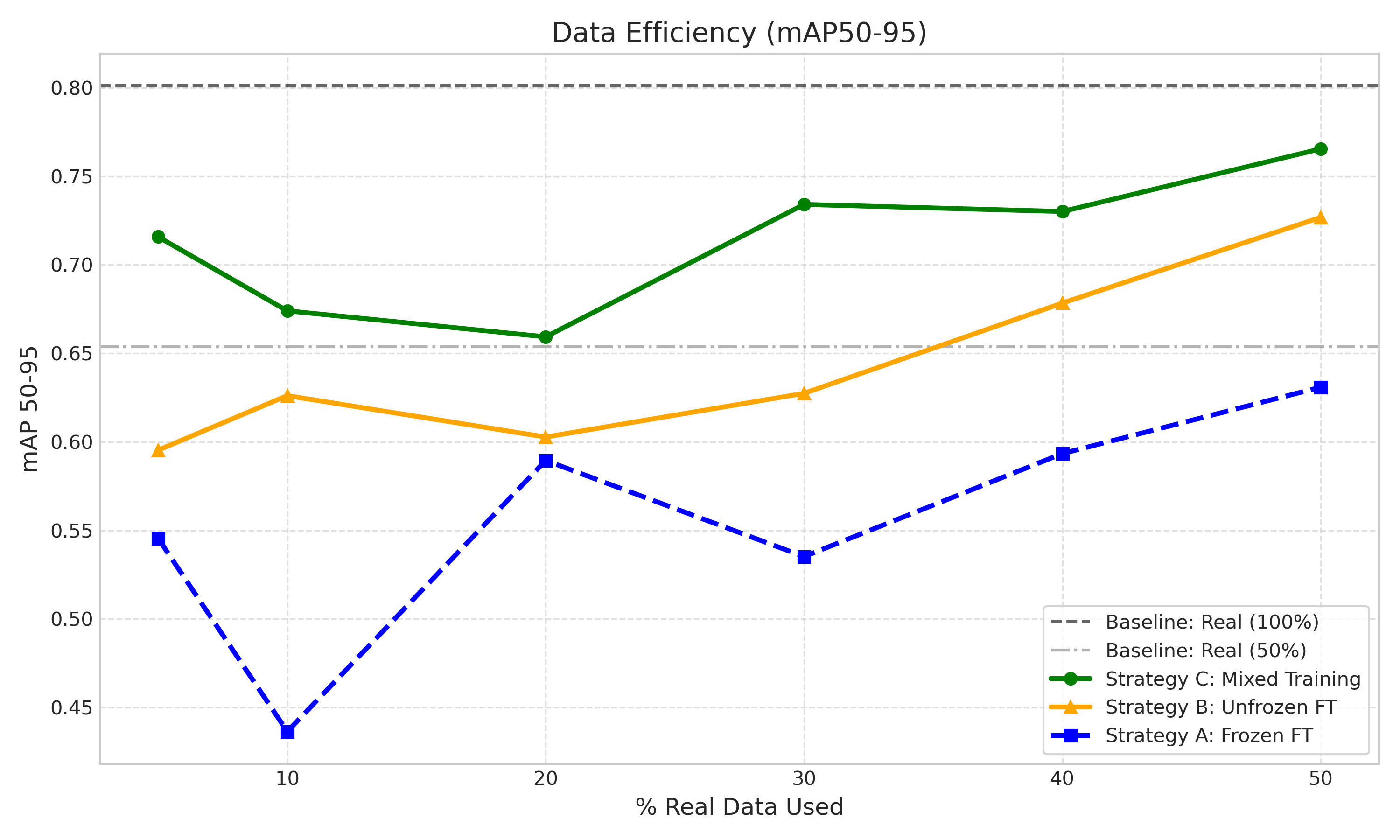}
    \caption{Data Efficiency Analysis. \textbf{Mixed Training} (Green) consistently outperforms both fine-tuning strategies. Notably, unfreezing the backbone (Strategy B, Orange) provides a substantial gain over linear probing (Strategy A, Blue), particularly as more real data becomes available.}
    \label{fig:efficiency_plot}
\end{figure}

\subsection{Isolation of Synthetic Data Contribution}
To verify the specific contribution of synthetic data, Table \ref{tab:ablation} compares the best performing synthetic strategy (Mixed Training) against a baseline trained exclusively on real data at the 40\% and 50\% benchmarks.

The ablation confirms that the gains from Strategy C are attributable to the complementary information introduced by synthetic data. At the 40\% regime, Mixed Training improves both localization quality \textbf{mAP50-95} and the operational metrics \textbf{Precision/Recall} compared to the Real-Only baseline, indicating fewer missed detections and false positives in a counting-oriented setting. At the 50\% regime, the gap narrows: Full Fine-Tuning (Strategy B) becomes highly competitive in \textbf{Recall}, suggesting that once sufficient real data is available, adapting the entire backbone can recover much of the remaining domain mismatch. Nevertheless, Mixed Training remains strong, especially in \textbf{mAP50-95} and \textbf{Precision} supporting the conclusion that synthetic data can reduce the amount of real data needed to reach high and stable performance.
 
\begin{table}[htbp]
\caption{Ablation Study: Impact of Synthetic Data at 40\% and 50\% Regimes}
\label{tab:ablation}
\begin{center}
\resizebox{\columnwidth}{!}{%
\begin{tabular}{lcccc}
\toprule
\textbf{Configuration} & \textbf{mAP50} & \textbf{mAP50-95} & \textbf{Precision} & \textbf{Recall} \\
\midrule
\multicolumn{5}{l}{\textit{At 40\% Real Data}} \\
Real-Only Baseline & 0.9279 & 0.6902 & 0.8836 & 0.8457 \\
Strategy C (Mixed) & \textbf{0.9402} & \textbf{0.7301} & \textbf{0.9141} & \textbf{0.8916} \\
\midrule
\multicolumn{5}{l}{\textit{At 50\% Real Data}} \\
Real-Only Baseline & 0.9360 & 0.6536 & 0.8817 & 0.8825 \\
Strategy B (Unfrozen) & 0.9430 & 0.7268 & 0.8890 & \textbf{0.9110} \\
Strategy C (Mixed) & 0.9366 & \textbf{0.7655} & \textbf{0.9157} & 0.8846 \\
\bottomrule
\end{tabular}}
\end{center}
\end{table}

\begin{figure}[htbp]
    \centering
    \begin{subfigure}[b]{0.48\textwidth}
        \centering
        \includegraphics[width=\textwidth]{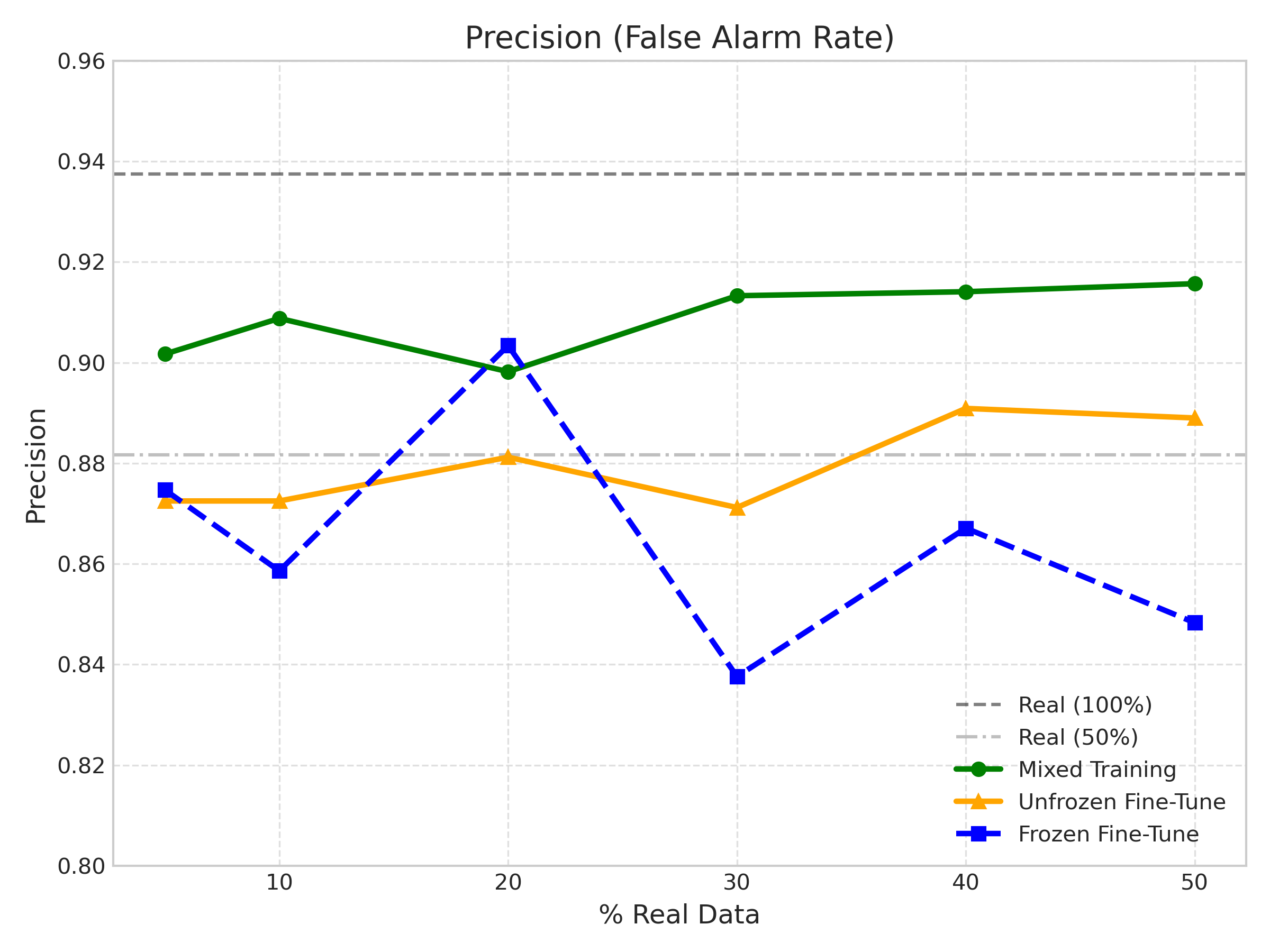}
        \caption{Precision (False Alarm Rate)}
        \label{fig:precision}
    \end{subfigure}
    \hfill 
    \begin{subfigure}[b]{0.48\textwidth}
        \centering
        \includegraphics[width=\textwidth]{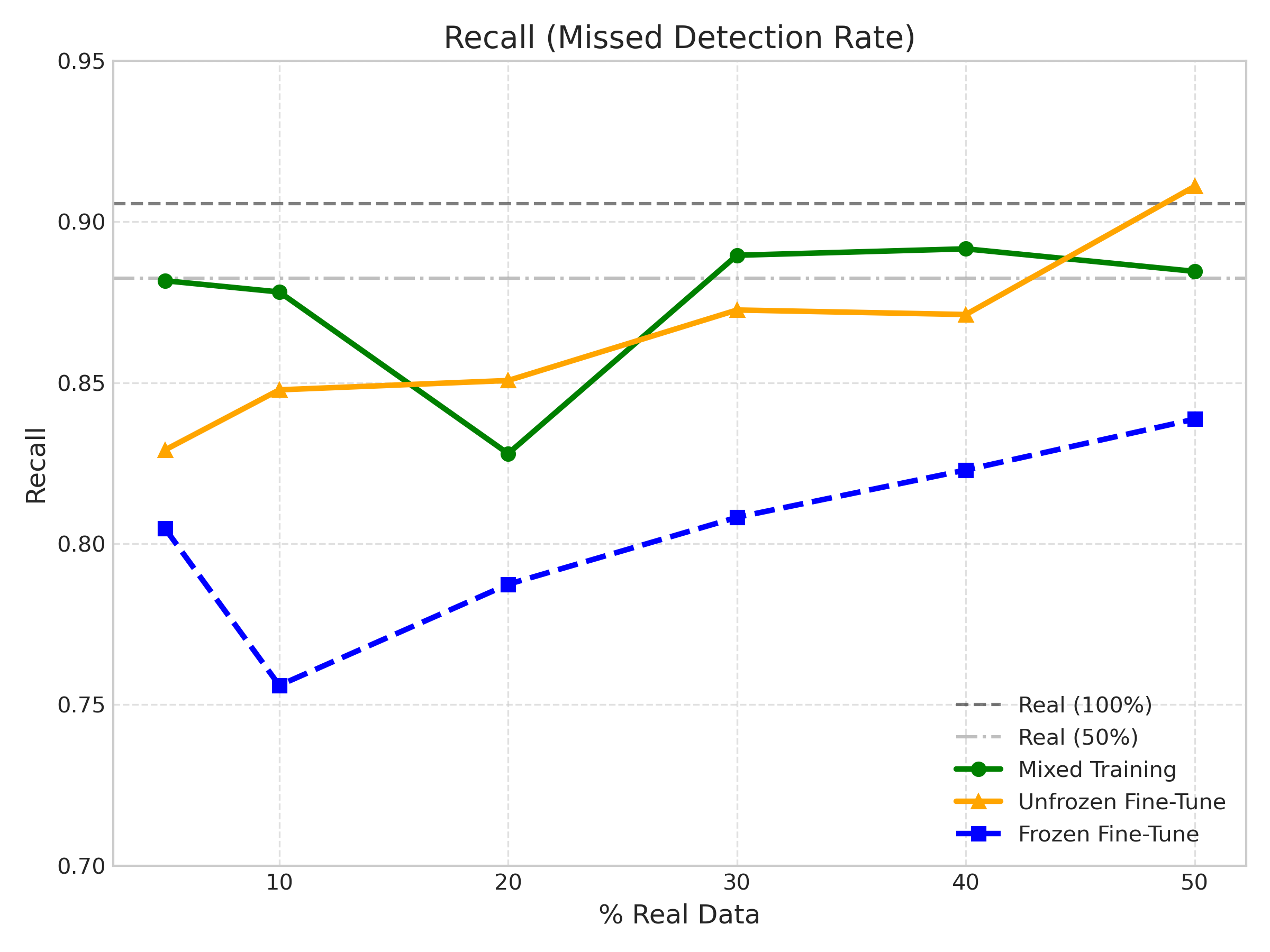}
        \caption{Recall (Missed Detection Rate)}
        \label{fig:recall}
    \end{subfigure}
    
    \caption{Operational Metrics Analysis. \textbf{(a)} Precision Comparison: All strategies maintain high precision, with Mixed Training (Green) consistently leading. \textbf{(b)} Recall Comparison: Strategy B (Orange) shows a strong recovery in Recall at 50\% data, surpassing even the Mixed Strategy, suggesting that full fine-tuning becomes highly effective once sufficient real data is available to adapt the weights.}
    \label{fig:prec_recall_combined}
\end{figure}

\subsection{Multi-Seed Validation and Training Stability}
To ensure that the observed performance gains are robust and not merely the result of a favorable random initialization or a specific data split, we conducted a multi-seed validation study. The models were independently trained across three distinct random seeds (\textit{seeds: 42, 123, 456}) and evaluated on the held-out test set. Table \ref{tab:stability} reports the mean performance alongside the standard deviation ($\pm$) for each metric.

\begin{table}[htbp]
\caption{Multi-Seed Validation Results (Mean $\pm$ Standard Deviation across 3 seeds)}
\label{tab:stability}
\begin{center}
\resizebox{\columnwidth}{!}{%
\begin{tabular}{lccccc}
\toprule
\textbf{Strategy} & \textbf{Data \%} & \textbf{mAP50} & \textbf{mAP50-95} & \textbf{Precision} & \textbf{Recall} \\
\midrule
\multicolumn{6}{c}{\textit{Real-Only Baseline}} \\
Real-Only & 10\% & 0.7663 $\pm$ 0.0000 & 0.4681 $\pm$ 0.0000 & 0.7797 $\pm$ 0.0000 & 0.6489 $\pm$ 0.0000 \\
Real-Only & 20\% & 0.8942 $\pm$ 0.0156 & 0.6287 $\pm$ 0.0578 & 0.8649 $\pm$ 0.0003 & 0.8164 $\pm$ 0.0302 \\
\midrule
\multicolumn{6}{c}{\textit{Strategy A: Linear Probing (Frozen)}} \\
Fine-Tuning Frozen Backbone  & 5\%  & 0.8653 $\pm$ 0.0413 & 0.5808 $\pm$ 0.0490 & 0.8147 $\pm$ 0.0562 & 0.7695 $\pm$ 0.0476 \\
Fine-Tuning Frozen Backbone  & 10\% & 0.8811 $\pm$ 0.0248 & 0.5386 $\pm$ 0.0724 & 0.8395 $\pm$ 0.0295 & 0.8033 $\pm$ 0.0354 \\
Fine-Tuning Frozen Backbone  & 20\% & 0.8985 $\pm$ 0.0112 & 0.6048 $\pm$ 0.0045 & 0.8563 $\pm$ 0.0062 & 0.8114 $\pm$ 0.0158 \\
Fine-Tuning Frozen Backbone  & 30\% & 0.9145 $\pm$ 0.0198 & 0.6461 $\pm$ 0.0475 & 0.8786 $\pm$ 0.0150 & 0.8351 $\pm$ 0.0236 \\
Fine-Tuning Frozen Backbone  & 40\% & 0.8965 $\pm$ 0.0179 & 0.6138 $\pm$ 0.0313 & 0.8731 $\pm$ 0.0115 & 0.8081 $\pm$ 0.0213 \\
\midrule
\multicolumn{6}{c}{\textit{Strategy B: Full Fine-Tuning (Unfrozen)}} \\
Full Fine-Tuning & 5\%  & 0.9228 $\pm$ 0.0121 & 0.6352 $\pm$ 0.0341 & 0.8888 $\pm$ 0.0131 & 0.8392 $\pm$ 0.0132 \\
Full Fine-Tuning & 10\% & 0.9274 $\pm$ 0.0221 & 0.6405 $\pm$ 0.0243 & 0.8672 $\pm$ 0.0389 & 0.8648 $\pm$ 0.0305 \\
Full Fine-Tuning & 20\% & 0.9220 $\pm$ 0.0000 & 0.6009 $\pm$ 0.0000 & 0.8695 $\pm$ 0.0000 & 0.8653 $\pm$ 0.0000 \\
Full Fine-Tuning & 30\% & 0.9445 $\pm$ 0.0187 & 0.6626 $\pm$ 0.0417 & 0.8899 $\pm$ 0.0185 & 0.8940 $\pm$ 0.0215 \\
\midrule
\multicolumn{6}{c}{\textit{Strategy C: Mixed Training}} \\
Mixed Training & 5\%  & 0.9285 $\pm$ 0.0099 & 0.6911 $\pm$ 0.0368 & 0.9056 $\pm$ 0.0112 & 0.8551 $\pm$ 0.0208 \\
Mixed Training & 10\% & 0.8594 $\pm$ 0.0912 & 0.6009 $\pm$ 0.1238 & 0.8698 $\pm$ 0.0450 & 0.7653 $\pm$ 0.1229 \\
Mixed Training & 20\% & 0.9272 $\pm$ 0.0166 & 0.6793 $\pm$ 0.0386 & 0.9068 $\pm$ 0.0226 & 0.8483 $\pm$ 0.0262 \\
Mixed Training & 30\% & 0.9333 $\pm$ 0.0181 & 0.7103 $\pm$ 0.0309 & 0.8923 $\pm$ 0.0294 & 0.8685 $\pm$ 0.0139 \\
Mixed Training & 40\% & 0.9474 $\pm$ 0.0047 & 0.7483 $\pm$ 0.0341 & 0.9236 $\pm$ 0.0018 & 0.8893 $\pm$ 0.0026 \\
\bottomrule
\end{tabular}}
\end{center}
\end{table}

The results reveal a high degree of training stability across all methodologies. For the majority of the experiments, the standard deviation remains exceptionally low (e.g., $< \pm 0.03$ for mAP50), confirming that the network consistently converges to a similar local minimum regardless of the random seed. 

Notably, \textbf{Strategy C (Mixed Training)} exhibits remarkable stability even at the extremes of the data scale. At the highly data-scarce 5\% regime, variance remains minimal (mAP50 $\pm$ 0.0099), demonstrating that the strong geometric prior provided by the synthetic data prevents the model from overfitting to specific real-world data shuffles. As the real data fraction increases to 40\%, the variance tightens even further (mAP50 $\pm$ 0.0047, Precision $\pm$ 0.0018). While a slight variance spike is observed at the 10\% Mixed benchmark, it resolves rapidly by the 20\% mark. Overall, this multi-seed validation confirms that integrating synthetic OBB data into the training pipeline is a highly reliable and reproducible strategy for real-world deployment.

\subsection{Qualitative Results}
\begin{figure}[htbp]
    \centering
    \subfloat[Baseline (50\% Real-Only): Missed detections in the nested chain and a double detection of the first trolley.\label{fig:qual_baseline_50}]{%
        \includegraphics[width=0.85\columnwidth]{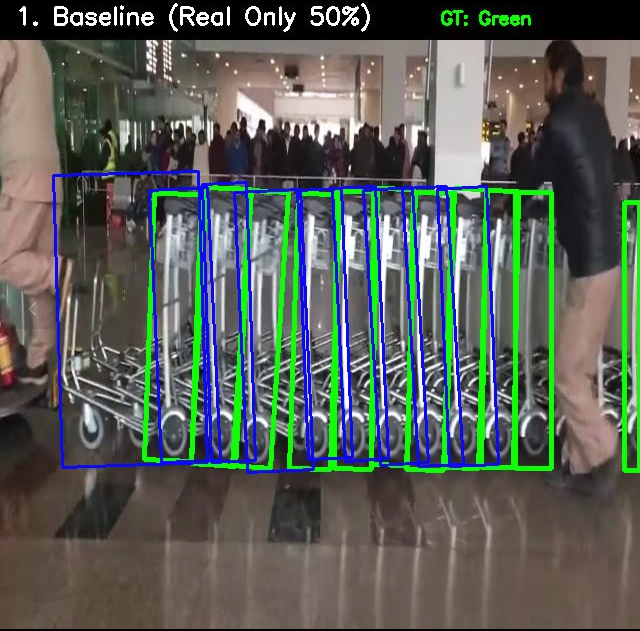}%
    }
    
    \par\smallskip
    \subfloat[Fine-tuned (all layers, 50\% real): Double detection of the first trolley.\label{fig:qual_finetune_50}]{%
        \includegraphics[width=0.85\columnwidth]{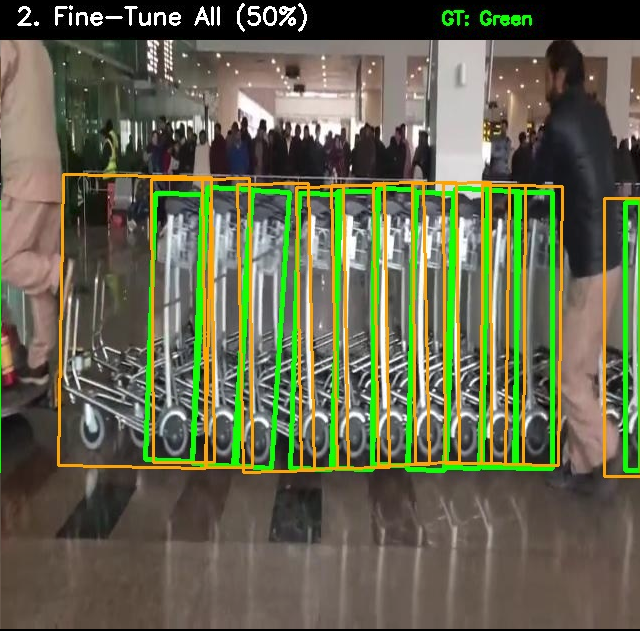}%
    }

    \par\smallskip
    \subfloat[Mixed (40\% Real + Syn): Correctly detects the nested object.\label{fig:qual_mixed}]{%
        \includegraphics[width=0.85\columnwidth]{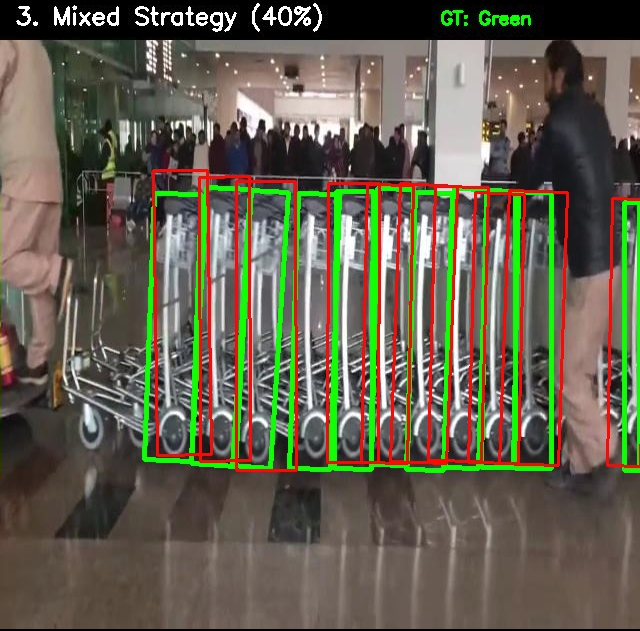}%
    }
    \caption{Qualitative comparison on a challenging test frame.}\phantomsection
    \label{fig:detections}
\end{figure}

Figure \ref{fig:detections} illustrates the qualitative performance of the Mixed Model (40\%) compared to the Real-Only Baseline (40\%).

Beyond the aggregate metrics, these examples highlight the failure modes that most directly impact trolley counting. In the real-only baselines, dense ``chained'' formations often lead to \textit{missed detections} of deeply nested units (reduced Recall) and occasional \textit{duplicate detections} on salient parts such as the leading trolley or handle structure (reduced Precision). In contrast, the mixed strategy benefits from the synthetic prior learned on highly crowded configurations, producing more consistent instance separation in overlapping chains and improving robustness to cluttered backgrounds. Overall, the qualitative results align with the quantitative trends: synthetic data primarily contributes by stabilizing detections in the most congested scenes, where counting errors are otherwise dominated by occlusions and heavy inter-object overlap.

\section{Discussion and Analysis}
\label{sec:discussion}

\IEEEPARstart{T}{his} section interprets the experimental results to understand \textit{when} and \textit{why} synthetic data contributes to performance, and analyzes the trade offs between the three distinct training strategies.

\subsection{The "Data Efficiency" Multiplier}
Our central finding is that synthetic data acts as a powerful regularizer when real world data is scarce. As illustrated in the Data Efficiency Plot (Fig. \ref{fig:efficiency_plot}), \textbf{Strategy C (Mixed Training)} consistently provides the highest Mean Average Precision (mAP) in low data regimes (5\% to 30\%).

\begin{itemize}
    \item \textbf{Quantifying the Gain:} The Mixed Model achieves \textbf{0.73 mAP} using only \textbf{30\%} of the available real data. This surpasses the performance of the Real Only baseline trained on \textbf{40\%} data (0.69 mAP).
    \item \textbf{Operational Implication:} This implies a \textbf{25\% reduction in annotation effort} (30\% vs 40\%) to achieve equivalent detection performance. For large scale airport deployments, this efficiency translates to significantly lower setup costs.
\end{itemize}

\subsection{Backbone Plasticity: Why Linear Probing Fails}
The poor performance of \textbf{Strategy A (Linear Probing)}, where the backbone layers are frozen, reveals a critical insight into the Simulation to Reality gap. 
The fact that Strategy A performs significantly worse than Strategy B (Full Fine Tuning) indicates that the \textit{features} learned in the synthetic domain (e.g., textureless edges, perfect lighting) are not directly transferable to the real world. The backbone \textit{must} be allowed to update its weights to adapt to the specific noise and texture of the airport CCTV footage. Freezing the network prevents this necessary adaptation, resulting in a model that cannot generalize well despite having a tuned prediction head.

\subsection{The Trade off: Catastrophic Forgetting vs. Adaptation}
The comparison between \textbf{Strategy B (Full Fine Tuning)} and \textbf{Strategy C (Mixed Training)} highlights the balance between learning new domains and retaining old priors.

\begin{enumerate}
    \item \textbf{Low Data Regimes (5-30\%):} Mixed Training significantly outperforms Full Fine Tuning. In this phase, Strategy B suffers from overfitting; the small real world subset is insufficient to update the entire network without degrading the general features. Strategy C avoids this because the synthetic data acts as an "anchor," preserving the geometric understanding of trolleys while the real data slowly updates the texture representations.
    
    \item \textbf{High Data Regimes (50\%+):} A crossover occurs at the 50\% mark. As seen in Figure \ref{fig:prec_recall_combined}b, Strategy B achieves the highest \textbf{Recall (91.1\%)}, surpassing even the Mixed Strategy. At this level of real supervision, unfreezing and finetuning all layers generalizes better to the real test distribution especially in terms of Recall because the backbone can fully adapt to real world texture, noise, and viewpoint statistics. Consequently, Strategy B becomes more competitive with the Mixed Training approach (and slightly stronger in Recall at 50\%), while Mixed Training remains generally the more reliable choice specially in low data regims.
\end{enumerate}

\subsection{Bridging the Domain Gap}
The extremely low performance of the \textit{Synthetic Only} baseline (0.21 mAP) confirms a substantial domain gap. The synthetic model captures the \textit{geometry} of the trolleys (hence the reasonable detection of shapes) but fails to generalize to the specific \textit{texture} and \textit{lighting} of the airport environment.
However, the results show that this gap is bridgeable. The synthetic data provides the "structural scaffolding" (learning to separate nested objects), while the real data provides the "texture mapping." This synergy is best utilized in Strategy C, which allows the model to detect complex occlusion cases (as seen in Fig. \ref{fig:qual_mixed}) that the Real Only model misses due to the rarity of such dense configurations in the small real training set.


\section{Limitations and Future Work}
\label{sec:limitations}

While this study demonstrates the significant potential of synthetic data for airport asset detection, several limitations must be acknowledged. 

\subsection{Current Limitations}
\begin{itemize}
    \item \textbf{Asset and Environment Diversity:} The current Digital Twin is tailored to the architectural layout and lighting conditions of Algiers International Airport, and the synthetic data includes only the two trolley variants used at this site. As a result, the extent of \textit{zero-shot} generalization to airports with different illumination profiles (e.g., strong sunlight variations) or substantially different trolley designs (e.g., three-wheeled or locally customized models) remains unverified.
    \item \textbf{Synthetic Realism Constraints:} As indicated by our failure case analysis, the residual \textit{simulation-to-reality} gap is dominated by appearance factors (texture, illumination, and specular effects) rather than by geometric mismatch. The current rendering pipeline does not fully model high-fidelity Physically Based Rendering (PBR) phenomena (e.g., accurate ray-traced reflections on metallic surfaces and lens glare), which likely limits the benefit attainable from synthetic pretraining alone.
\end{itemize}

\subsection{Future Directions}
Addressing these constraints opens several promising avenues for future research:
\begin{itemize}
    \item \textbf{Unsupervised Domain Adaptation (UDA):} Future work will explore UDA techniques (e.g., adversarial feature alignment) to bridge the texture gap without requiring \textit{any} manual labeling of the target real-world domain.
    \item \textbf{Expansion to Other Airport Assets:} The Digital Twin methodology established here serves as a blueprint. Future iterations will expand the synthetic asset library to include wheelchairs, cargo loaders, and autonomous cleaning robots, moving toward a holistic "Smart Airport" logistical monitoring system.
\end{itemize}
\section{Data and Code Availability}
\label{sec:codes}

To support reproducibility, we provide the full experimental codebase in our public GitHub repository: \textbf{\href{https://github.com/djallilou13/yolo-obb-experiments}{https://github.com/djallilou13/yolo-obb-experiments}}. We also provide the project website: \textbf{\href{https://airport-trolley-dataset.vercel.app/}{https://airport-trolley-dataset.vercel.app/}}, where readers can find extended documentation and access the dataset used in this paper.
\section{Conclusion}
\label{sec:conclusion}

\IEEEPARstart{T}{he} development of automated logistical monitoring systems in airports is frequently bottlenecked by the high cost, privacy concerns, and security restrictions associated with large scale operational video data collection. This study proposed a data efficient framework to overcome these barriers by leveraging a synthetic "Digital Twin" of the Algiers International Airport. 

We generated a highly dense, automatically annotated synthetic dataset of oriented bounding boxes (OBB) and rigorously evaluated its utility across three adaptation strategies: Linear Probing, Full Finetuning, and Mixed Training. 

Our findings demonstrate that synthetic data acts as a highly effective regularizer in data scarce environments. Specifically, the \textbf{Mixed Training} strategy reduced the requirement for labeled real world data by 35\% while achieving superior data efficiency (0.73 mAP at a 30\% data split). 

Moving forward, several promising avenues exist for extending this research and paving the way for future publications. First, the "Simulation to Reality" pipeline can be further advanced by integrating Unsupervised Domain Adaptation (UDA) frameworks such as adversarial feature alignment or CycleGAN based image to image translation to theoretically eliminate the need for real world annotations entirely. Additionally, researching the quantization and optimization of the YOLO OBB models for deployment on low power edge computing devices (e.g., NVIDIA Jetson) would provide a critical step toward realizing a decentralized, privacy preserving, real time monitoring system.

Ultimately, this research contributes a scalable, semi automated methodology for visual asset detection. By significantly reducing the manual annotation burden while maintaining strict operational metrics (high Precision and Recall), this approach provides a practical pathway for deploying robust computer vision systems in complex, highly regulated logistical environments.

\bibliographystyle{IEEEtran}
 \bibliography{references} 

\end{document}